\pdfoutput=1

\documentclass[11pt]{article}

\usepackage[final]{acl}

\usepackage{times}
\usepackage{latexsym}
\usepackage{graphicx}
\usepackage{amsmath}
\usepackage{comment}
\usepackage{booktabs}
\usepackage[T1]{fontenc}

\usepackage[utf8]{inputenc}

\usepackage{booktabs}
\usepackage[table]{xcolor}
\usepackage{multirow}
\definecolor{rowgray}{gray}{0.98}
\usepackage{inconsolata}
\usepackage{multirow}
\usepackage{placeins}
\usepackage[table,xcdraw]{xcolor}
\usepackage{pifont}
\usepackage{graphicx}
\usepackage[inkscapelatex=false]{svg}
\usepackage{hyperref}
\usepackage{arydshln}
\usepackage{footmisc} 
\title{Evaluating Customized vs. Generalist Transformer-based Models for Legal Contract Classification}

\author{
  Amrita Singh, 
  H. Suhan Karaca, 
  Aditya Joshi, 
  Hye-young Paik, 
  Jiaojiao Jiang \\
  School of Computer Science and Engineering \\
  University of New South Wales (UNSW), Sydney \\
  \footnotetext[1]{These authors contributed equally to this work.}
}

\usepackage{ amssymb}
\newcommand{\xmark}{\ding{53}} 
\begin{document}
\maketitle
\begin{abstract}
Despite advances in legal NLP, no comprehensive evaluation of Transformer-based models customized for legal tasks (referred to as `legal-specific' models in this paper) exists for contract classification tasks. To address this gap, we present an evaluation of 13 legal-specific transformer-based models on 3 English-language contract classification tasks and compare them with 9 generalist models. The results show that legal-specific models consistently outperform generalist models, especially on tasks requiring nuanced legal understanding. They also help reduce misclassification of rare classes in imbalanced datasets. Legal-BERT and Contracts-BERT establish new SOTAs on two of the three tasks, despite having 69\% fewer parameters than the best-performing generalist models.  We also identify CaseLaw-BERT and LexLM as strong additional baselines for contract classification. Our results highlight the shortcomings of generalist models, emphasizing the need for domain-specific customization, particularly in the context of legal applications.
\end{abstract}

\section{Introduction}
\label{Introduction}
\begin{figure}[ht!]
\centering
\includegraphics[width=0.4\textwidth]{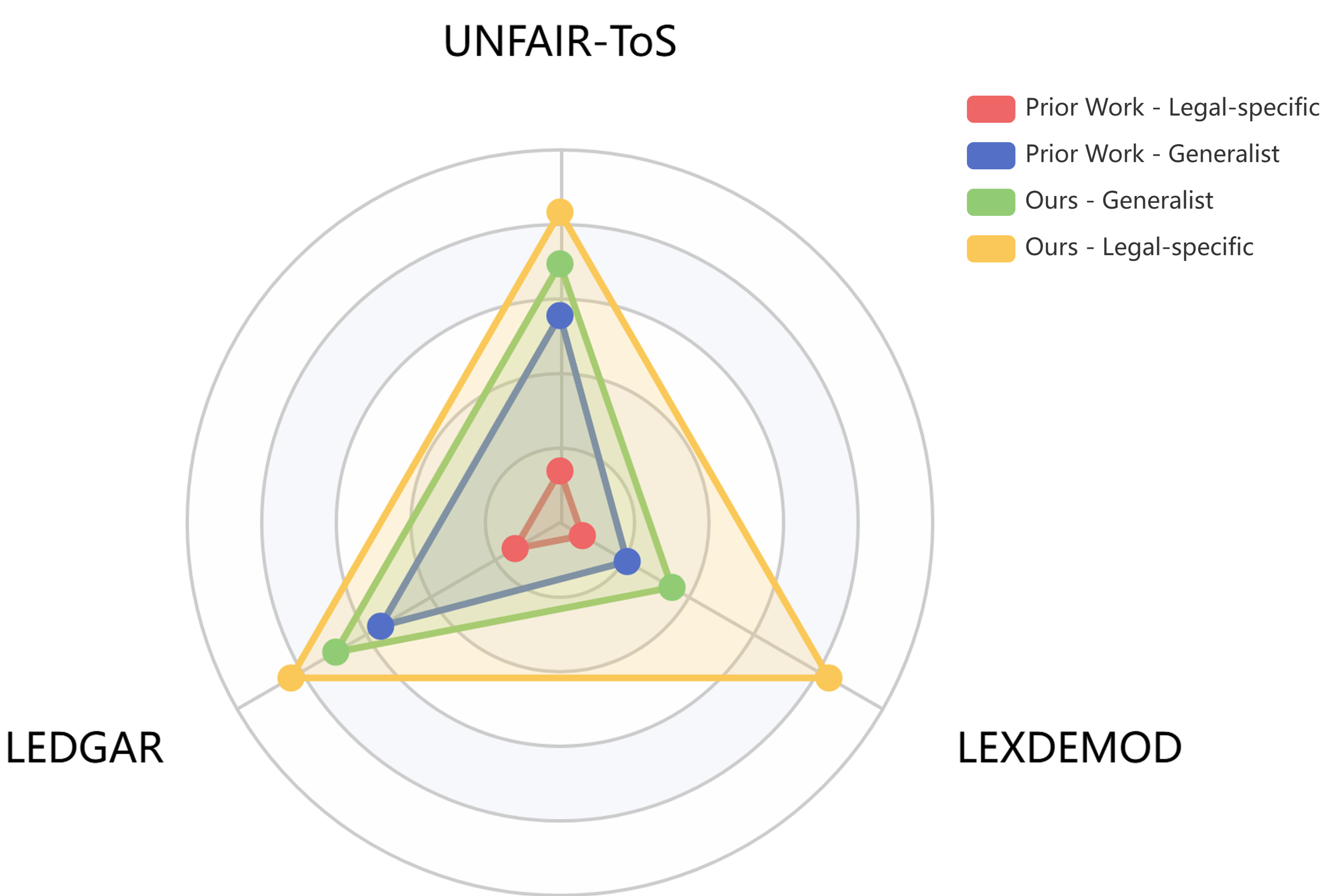}
\caption{Comparison of our legal-specific models evaluation and coverage with prior work.}
\label{Figure1}
\end{figure}

Recent work suggests that open-source legal-specific models offer a promising, cost-effective, and privacy-preserving alternative to generalist models \cite{singh2025survey,bhambhoria2024evaluating, chalkidis2020legal}. However, despite their advantages, these models remain significantly underutilized in current legal contract classification tasks, as noted by \citet{singh2025survey}. As illustrated in Figure \ref{Figure1}, legal-specific models are rarely evaluated, and mostly generalist models are preferred in prior work on three popular and freely available contract classification tasks: Unfair Contractual Terms Identification \cite{lippi2019claudette, chalkidis2022lexglue}, Contractual Provision Topic Classification \cite{tuggener2020ledgar, chalkidis2022lexglue}, and Agent-Specific Deontic Modality Detection \cite{sancheti2022agent}.
\\
Despite the legal nature of documents/tasks, researchers have continued to favor generalist models over legal-specific models. This holds true not only for decoder models (which is the predominant context in which `customization' is used) but also for encoder models that must be adopted to the legal domain. In some cases, legal-specific models are excluded entirely. For instance, recent studies such as \citet{guha2023legalbench} and \citet{singh2024data}, which explicitly focus on legal downstream tasks, do not include any legal-specific models in their benchmarking evaluations.
Therefore, this paper addresses the Research Question \textbf{(RQ):} \emph{How do legal-specific models perform compared to generalist models on legal contract classification tasks?}
To address this question, we present a comprehensive evaluation of 13 open-source legal-specific models with 9 generalist models across the three distinct contract classification tasks. Our results reveal consistent improvements in performance for legal-specific models, particularly on tasks where legal and domain-specific semantics are critical. This benchmark serves as a resource for the community, offering a clearer understanding of model suitability and performance across tasks and model types. The contributions of this work are as follows: (a) \emph{To the best of our knowledge, we present the first benchmarking of  multiple legal-specific models across multiple contract classification tasks}; (b) We systematically compare their performance with that of generalist models; (c) We identify model strengths, weaknesses, and task-specific challenges, offering insights for future research and deployment.
\section{Contract Classification Tasks and Datasets}
\label{Section3}
To evaluate the effectiveness of models customized for the legal setting vis-a-vis generalist models, we consider the following factors to select the datasets and tasks. We select the \textbf{language} to be English due to the availability of datasets and models. \textbf{Publicly available}, well-documented datasets are used, each large enough for stable training and evaluation. Proprietary, non-public, and very small datasets (under 3K sentences) are avoided to ensure reproducibility and generalizability. This criterion modifies and adapts the selection guidelines of \citet{chalkidis2022lexglue}. Datasets are chosen where SOTA generalist language models do not achieve near-perfect performance \cite{lippi2019claudette, tuggener2020ledgar, sancheti2022agent}, ensuring that benchmarking legal-specific language models remains \textbf{challenging}.
 For the chosen tasks to reflect \textbf{relevance and diversity}, we select tasks that test a model’s understanding of legal language, structure, and semantics. As shown below, three distinct tasks are selected, each using a different dataset and representing a unique contract classification scenario in terms of dataset size and number of classes.

Consequently, the following datasets are selected:
\textbf{UNFAIR-ToS:} The UNFAIR-ToS dataset (\textbf{train (5.5k), dev (2.3k), and test (1.6k)}) from \citet{chalkidis2022lexglue} is a \textbf{multi-label classification} dataset that is used to identify unfair contractual terms in Terms of Service (ToS) documents from online platforms like YouTube. Each sentence is annotated with one or more of \textbf{8 unfairness categories, plus 1 unlabeled class} for sentences that do not indicate any potential violation of European consumer law \citet{eu-directive-93-13}.

\textbf{LEDGAR:} The LEDGAR dataset (\textbf{train (60k), dev (10k), and test (10k)}) from \citet{chalkidis2022lexglue} is used to classify the principal topic of provisions in Exhibit 10 material contracts (e.g., employment, lease, non-disclosure) filed with the US Securities and Exchange Commission (SEC) via \citet{sec-edgar}. Each provision (paragraph) is labeled with one of \textbf{100 contract topics}, making it a \textbf{multi-class classification} task.

\textbf{LEXDEMOD:} The LEXDEMOD dataset (\textbf{train (4.2k), dev (330), and test (1.7k)}) from \citet{sancheti2022agent} detects deontic modality in agent-based contract clauses from lease agreements sourced from the LEDGAR dataset. Each clause (sentence) is annotated with one or more of \textbf{6 deontic modality types plus 1 none class} , making it a \textbf{multi-label classification} task. Labels are linked to an agent (party) in the sentence, representing their deontic status (e.g., Obligation, Entitlement, Prohibition). \emph{The train/dev/test split is as reported in the original paper}. Statistics and examples for all datasets are in Appendix \ref{AppendixA} while the class distribution is in Figure \ref{Appendix_Figure0}.
\begin{table*}[ht!]
\centering
\resizebox{0.94\textwidth}{!}{
\begin{tabular}{lclcllclll}
\toprule
 & & \multicolumn{1}{c}{} & \multicolumn{1}{c}{} & \multicolumn{2}{c}{\textbf{UNFAIR-ToS}} & \multicolumn{2}{c}{\textbf{LEDGAR}} & \multicolumn{2}{c}{\textbf{LEXDEMOD}} \\
\cmidrule(lr){5-6} \cmidrule(lr){7-8} \cmidrule(lr){9-10}
\multirow{-2}{*}{} & \multirow{-2}{*}{\textbf{Method}} & \multicolumn{1}{c}{\multirow{-2}{*}{\textbf{Model}}} & \multicolumn{1}{c}{\multirow{-2}{*}{\textbf{\# Params}}} & \multicolumn{1}{c}{$\mu$-F1} & \multicolumn{1}{c}{m-F1} & \multicolumn{1}{c}{$\mu$-F1} & \multicolumn{1}{c}{m-F1} & \multicolumn{1}{c}{$\mu$-F1} & \multicolumn{1}{c}{m-F1} \\
\midrule
 & & BERT & 110M & 95.6 & 81.3 & 87.6 & 81.8 & - & 75.61 \\
 & & RoBERTa-base & 125M & 95.2 & 79.2 & 87.9 & 82.3 & - & 75.66 \\
 & & DeBERTa & 139M & 95.5 & 80.3 & 88.2 & 83.1 & - & - \\
 & & Longformer & 149M & 95.5 & 80.9 & 88.2 & 83.0 & - & - \\
 & & BigBird & 127M & 95.7 & 81.3 & 87.8 & 82.6 & - & - \\
\multirow{-8}{*}{\begin{tabular}[c]{@{}l@{}}Baselines reported from:\\
\cite{chalkidis2022lexglue}, \\
\cite{sancheti2022agent}\end{tabular}} & \multirow{-8}{*}{\begin{tabular}[c]{@{}c@{}}\\Generalist\\ Models \end{tabular}} & RoBERTa-large & 355M & 95.8 & \color[HTML]{3531FF}{81.6} & \color[HTML]{3531FF}{88.6} & \color[HTML]{3531FF}83.6 & - & \color[HTML]{3531FF}77.88 \\

\cite{biswas2025optimized} & Legal-KDD-Student & DistilBERT & 66M &  \color[HTML]{3531FF}{95.9} & 76.2 & 88.2 & 82.3 & - & - \\
\cite{shin2025improved} & GATs & Hierarchical BERT & 110M & 95.6 & 81.5 & 88.4 & 83.1 & - & - \\
\midrule
& & Llama-3.2 & 3B & 95.9 & 80.3 & 85 & 76.1 & 76.2 & 71.4 \\
\multirow{-11}{*}{} & \multirow{-2}{*}{\begin{tabular}[c]{@{}c@{}}Generalist\\ Models\end{tabular}} & Mistral & 7B & 96.0 & 80.7 & 86.4 & 79.2 & 76.0 & 71.2 \\
\cmidrule{2-10}
 & & Legal-BERT & 110M & 96.0 & 82.2 & \color[HTML]{FE0000}88.2 & \color[HTML]{FE0000}82.5 & \color[HTML]{FE0000}81.23 & \color[HTML]{FE0000}78.01 \\
 & & Contracts-BERT & 110M & 96.2 & \color[HTML]{FE0000}83.4 & 87.9 & 82.2 & 80.17 & 77.71 \\
 & & Legal-RoBERTa & 125M & 95.4 & 81.1 & 87.7 & 81.9 & 80.12 & 76.70 \\
 & & CaseLawBERT & 110M & 96.1 & 83.2 & 87.6 & 80.9 & 80.32 & 77.75 \\
 & & PoL-BERT & 340M & 94.6 & 77.9 & 86.0 & 79.1 & 41.35 & 15.75 \\
 & & InLegalBERT & 110M & 95.6 & 81.7 & 87.9 & 82.0 & 80.21 & 77.89 \\
 & & InCaseLawBERT & 110M & 95.5 & 81.1 & 87.5 & 82.1 & 79.16 & 76.83 \\
 & & CustomInLawBERT & 110M & 95.5 & 79.9 & 87.7 & 81.8 & 78.16 & 75.35 \\
 & & LexLM & 124M & 95.9 & 81.7 & 87.8 & 81.3 & 80.39 & 77.46 \\
 & & Legal-XLM-R & 184M & 94.9 & 78.2 & 87.7 & 81.7 & 80.62 & 77.56 \\
 & & LexT5 & 220M & 95.4 & 79.8 & 84.9 & 76.1 & 76.50 & 73.30 \\
 & & AdaptLLM & 7B & \color[HTML]{FE0000} 96.5 & 83.2 & 85.2 & 76.8 & 76.2 & 70.0 \\
\multirow{-15}{*}{Proposed} & \multirow{-11}{*}{\begin{tabular}[c]{@{}c@{}}Legal-specific\\ Models\end{tabular}} & SaulLM & 7B & 96.0 & 81.0 & 86.6 & 79.4 & 76.5 & 72.7 \\
\bottomrule
\end{tabular}
}
\caption{Performance of legal-specific and generalist models on three tasks: UNFAIR-ToS, LEDGAR, LEXDEMOD. Metrics: micro-F1 ($\mu$-F1) and macro-F1 (m-F1). Blue highlights the best generalist, and red highlights the best legal-specific model performance.}
\label{Table3}
\end{table*}
\begin{table}[ht!]
\centering
\resizebox{0.34\textwidth}{!}{
\begin{tabular}{l ll}
\toprule
\multicolumn{1}{c}{} &
  \multicolumn{2}{c}{\textbf{Mean $\pm$ Std}} \\
\cmidrule(lr){2-3}
\multicolumn{1}{c}{\multirow{-2}{*}{\textbf{\begin{tabular}[c]{@{}c@{}}Legal Specific\\ Models\end{tabular}}}} &
  $\mu$-F1 & m-F1 \\
\midrule
Legal-BERT &
  {\color[HTML]{FE0000} 88.48 ± 6.03} & {\color[HTML]{3531FF} 80.90 ± 2.05} \\
Contracts-BERT &
  {\color[HTML]{3531FF} 88.09 ± 6.55} & {\color[HTML]{FE0000} 81.10 ± 2.45} \\
Legal-RoBERTa &
  87.74 ± 6.24 & 79.90 ± 2.29\\
CaseLawBERT &
  88.01 ± 6.45 & {\color[HTML]{009901} 80.62 ± 2.23} \\
PoL-BERT &
  73.98 ± 23.34 & 57.58 ± 29.58 \\
InLegalBERT &
  87.90 ± 6.28 & 80.53 ± 1.87 \\
InCaseLawBERT &
  87.39 ± 6.67 & 80.01 ± 2.29 \\
CustomInLawBERT &
  87.12 ± 7.09 & 79.02 ± 2.71 \\
LexLM &
  {\color[HTML]{009901} 88.03 ± 6.33} & 80.15 ± 1.91 \\
Legal-XLM-R &
  87.74 ± 5.83 & 79.15 ± 1.82 \\
LexT5 &
    85.60 ± 7.73 & 76.40 ± 2.66  \\
AdaptLLM &
  85.97 ± 8.31 & 76.67 ± 5.39 \\
SaulLM &
  86.37 ± 7.96 & 77.70 ± 3.60 \\
\bottomrule
\end{tabular}
}
\caption{Aggregated scores (Mean $\pm$ Std) across three contract classification tasks. Red, blue, and green highlights indicate the first, second, and third best performances, respectively.}
\label{Table4}
\end{table}
\section{Experiment Setup}
\label{section4}
We perform task-specific fine-tuning using 13 legal-specific models on three datasets: LEDGAR, UNFAIR-ToS, and LEXDEMOD. We consider ten pre-trained encoder-based legal-specific models for fine-tuning. Nine of these are base-variant encoder models: Legal-BERT \cite{chalkidis2020legal}, Contracts-BERT \cite{chalkidis2020legal}, LegalRoBERTa \cite{geng2021legal}, CaseLaw-BERT \cite{zheng2021does}, InLegalBERT, InCaseLawBERT, and CustomInLawBERT \cite{paul-2022-pretraining}, Legal-XLM-R \cite{niklaus2024multilegalpile}, and LexLM \cite{chalkidis-garneau-etal-2023-lexlms}. One large-variant model, PoL-BERT \cite{henderson2022pile}, is included, as its base version is not present. We also evaluate two legal decoder-based models, AdaptLLM \cite{cheng2024adapting} and SaulLM-7B \cite{colombo2024saullm}, along with one legal encoder-decoder model, LexT5 \cite{t-y-s-s-etal-2024-lexsumm}, as it is the only model of its kind among legal-specific decoder-based and encoder-decoder models. Table \ref{Table2} in Appendix \ref{AppendixB} summarizes the key characteristics of each model, and a detailed description of them is provided in the same appendix. A detailed experimental setup is provided in Appendix \ref{AppendixC}. We compare these 13 legal-specific models with 9 generalist models. These include six base variant encoder models: BERT \cite{devlin2019bert}, RoBERTa-base \cite{liu2019roberta}, DeBERTa \cite{hedeberta}, Longformer \cite{beltagy2020longformer}, BigBird \cite{zaheer2020big}, and DistilBERT \cite{sanh2019distilbert}, along with one large variant, RoBERTa-large \cite{liu2019roberta}. We also compare two generalist decoder-based models, Llama-3.2 \cite{grattafiori2025llama} and Mistral \cite{jiang2023mistral7b}.
\section{Results and Analysis}
\textbf{Generalist models:} Table \ref{Table3} reports the test results of models across all three tasks, comparing legal-specific and generalist models. Among generalist models, RoBERTa-large (355M) outperforms others, including larger decoder-based models like Llama-3.2 (3B) and Mistral (7B). Figure \ref{Appendix_Figure0} shows that all datasets suffer from extreme long-tail class imbalance. Table \ref{Table3} confirms that decoder-based models perform poorly on rare classes, as reflected in their lower macro-F1 scores, while RoBERTa-large performs consistently better. This suggests that encoder-based models are more robust to extreme class imbalance than decoder-based models. Despite having fewer parameters, RoBERTa-large achieves higher micro-F1 and macro-F1, demonstrating that \textit{bidirectional context learning and discriminative fine-tuning are more crucial than model scale for legal classification tasks with long-tail distributions}.

\textbf{Legal-specific models:} Legal-specific models such as Contracts-BERT and Legal-BERT (110M) outperform RoBERTa-large on UNFAIR-ToS and LEXDEMOD, respectively, despite having 69\% fewer parameters. RoBERTa-large remains the best model for LEDGAR. Still, Legal-BERT delivers equivalent performance compared to generalist base variant models on this task, suggesting that both model size and task characteristics influence performance. The larger, legal-specific encoder-based models may be better suited for LEDGAR, as the dataset is very large. \emph{Overall, legal-specific base models deliver competitive performance and set new SOTAs on two of the three tasks, demonstrating the effectiveness of domain-specific pretraining, even at the base variant of models.}

\begin{figure}[ht!]
\centering
    \includegraphics[width=0.42\textwidth]{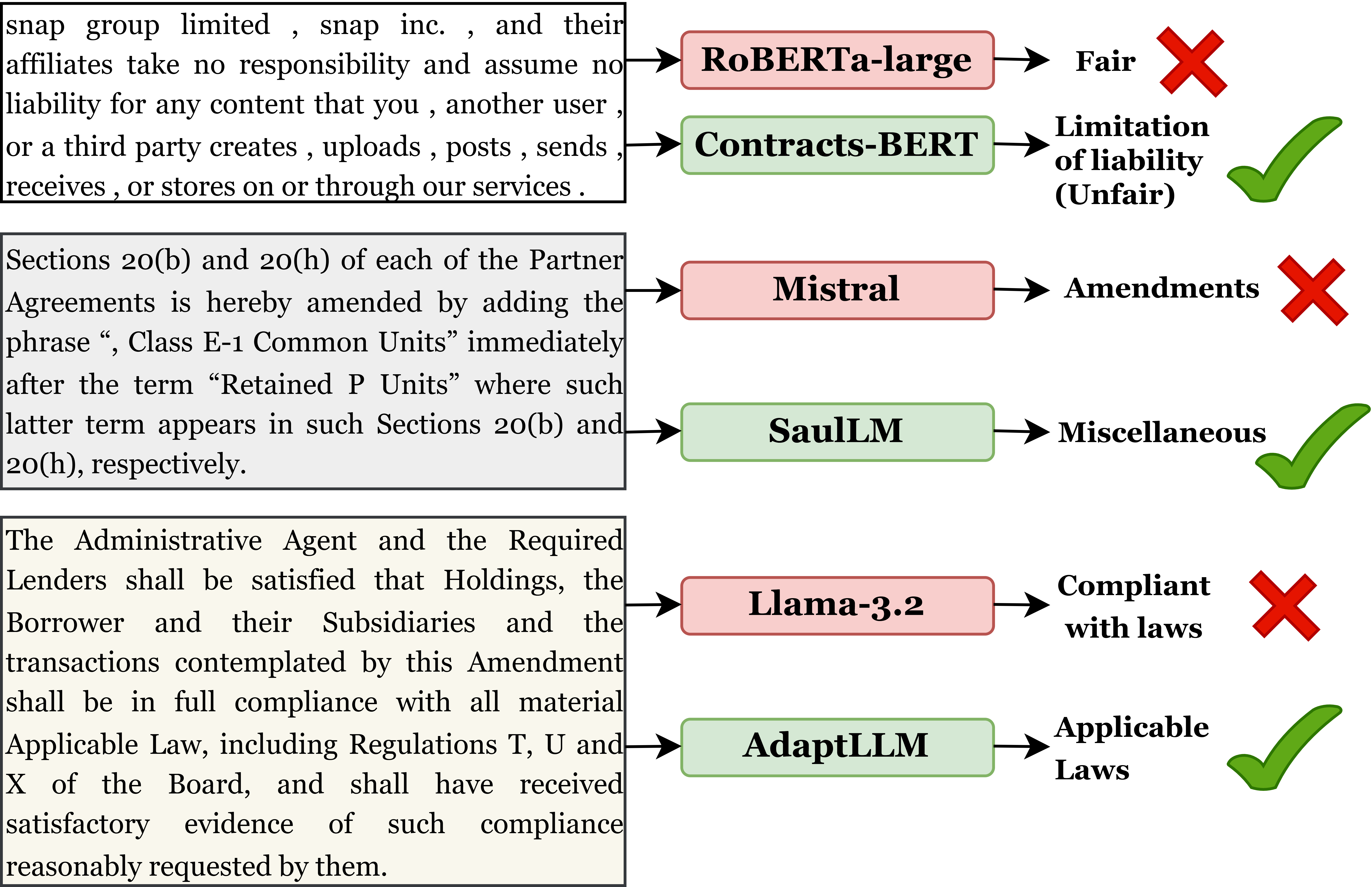}
    \caption{Example outputs showing customized models (green squares) correctly predict examples which were mis-classified by generalist models (red squares).}
    \label{Figure2}
\end{figure}

\textbf{Generalist models vs. Legal-specific models:} We further analyze the errors made by the best generalist model (RoBERTa-large) and examine whether the best legal-specific model (Contracts-BERT) corrects them on the Unfair-ToS dataset. As shown in Figure \ref{Figure2}, RoBERTa-large majorly misclassifies clauses belonging to rare categories, particularly \textit{Limitation of Liability} and \textit{Unilateral Termination}, whereas Contracts-BERT correctly classifies them, suggesting that domain-specific pretraining on contract corpora (refer Table \ref{Table2}) enables better representation of rare legal clause types that generalist models fail to capture. This analysis is further extended to decoder-based models by comparing generalist and their corresponding legal-specific variants on the LEDGAR dataset, as shown in Figure \ref{Figure2}. Specifically, Mistral and its legal-specific counterpart SaulLM, both sharing the same base architecture, are compared alongside Llama-3.2 and AdaptLLM, where AdaptLLM is built on Llama-1 but shares the same Llama architecture lineage. Across all cases, the results demonstrate that \textit{domain-specific pretraining helps in reducing the misclassification of rare classes in long-tail legal text classification}.

\textbf{Legal-specific models as baselines:} Table~\ref{Table4} presents aggregated test scores (Mean $\pm$ Std) across the three contract classification tasks. Despite class imbalance in all tasks, Legal-BERT achieves the highest aggregated $\mu$-F1, while Contracts-BERT leads in m-F1. Across both metrics, the top positions are consistently held by four legal-specific models: Legal-BERT, Contracts-BERT, CaseLaw-BERT, and LexLM. \emph{We conclude that these four models, Legal-BERT, Contracts-BERT, CaseLaw-BERT, and LexLM, should be considered strong baselines for contract classification tasks.}

\textbf{Limitations of Recent Legal-Specific Models:} Several recent legal-specific models, such as PoL-BERT, CustomInLawBERT, LexLM, Legal-XLM-R are pre-trained on large-scale legal corpora. Models like InLegalBERT and InCaseLawBERT are built on legal-specific base models rather than generalist models. However, older models like LegalBERT and ContractsBERT, pre-trained on just 354k and 76k legal documents (Table \ref{Table2}), still outperform recent base-variant legal-specific models (Table \ref{Table4}). This counterintuitive result can be attributed to a data distribution mismatch: more data only improves performance when it is in-distribution with the task. Recent models are trained on a broad mixture of legal genres, including court cases, legislation, and patents, which dilutes the contract-specific signal. Consequently, contracts remain underrepresented relative to other genres in these corpora. \textit{We conclude that future legal-specific models should incorporate a more diverse and representative set of contract documents, balanced alongside other legal genres, to improve performance on contract-based downstream tasks.}
\section{Conclusion}
This study benchmarks 13 legal-specific and 9 generalist models across three contract classification tasks. Encoder-based models outperform larger decoder-based models in long-tail legal classification, with domain-specific pretraining consistently improving rare class recognition. Legal-specific models set new state-of-the-art results on two of the three tasks, despite having 69\% fewer parameters than the best-performing generalist model. Legal-BERT, Contracts-BERT, CaseLaw-BERT, and LexLM serve as strong baselines for future research. Notably, older legal-specific models trained on smaller, more focused corpora outperform recent models trained on larger, genre-diverse corpora, highlighting that in-distribution pretraining is more critical than scale. Future work will explore retrieval-augmented approaches with legal-specific models to further enhance performance on rare and underrepresented legal clause types.

\section*{Limitations}
The limited availability of contract benchmark datasets in languages other than English presents a challenge for multilingual extension. As a result, this study focuses solely on English-language contract tasks, with evaluation on non-English data left for future research. This work also concentrates on the nuances of contract language and does not assess performance on other legal text types, such as statutes, court decisions, or legal opinions. Future research should broaden this evaluation to encompass a wider range of legal genres, recognizing that no single study can fully capture the entire legal domain. Additionally, this paper focuses on domain generalization alone while selecting the  legal domain.

\section*{Ethical Considerations}
This study uses only publicly available datasets, LEDGAR, UNFAIR-ToS, and LEXDEMOD, all of which contain contract clauses from publicly available contract documents. LEDGAR is derived from public U.S. SEC EDGAR filings, UNFAIR-ToS from company Terms of Service, and LEXDEMOD from lease clauses sourced from LEDGAR. This research does not offer legal advice, predict individual outcomes, or automate decisions affecting rights. It focuses solely on evaluating the performance of legal-specific models to inform future tools and research. While these models can support legal professionals, they are not substitutes for legal expertise. We acknowledge potential ethical risks if outputs are misused or inaccurate. By open-sourcing our evaluations, we aim to reduce reliance on proprietary tools, promote transparency, and expand access to legal AI research and development.

\bibliography{custom}
\appendix
\section{Dataset Statistics and Illustrative Examples}
\label{AppendixA}
\begin{figure*}[ht!]
    \includegraphics[width=0.8\textwidth]{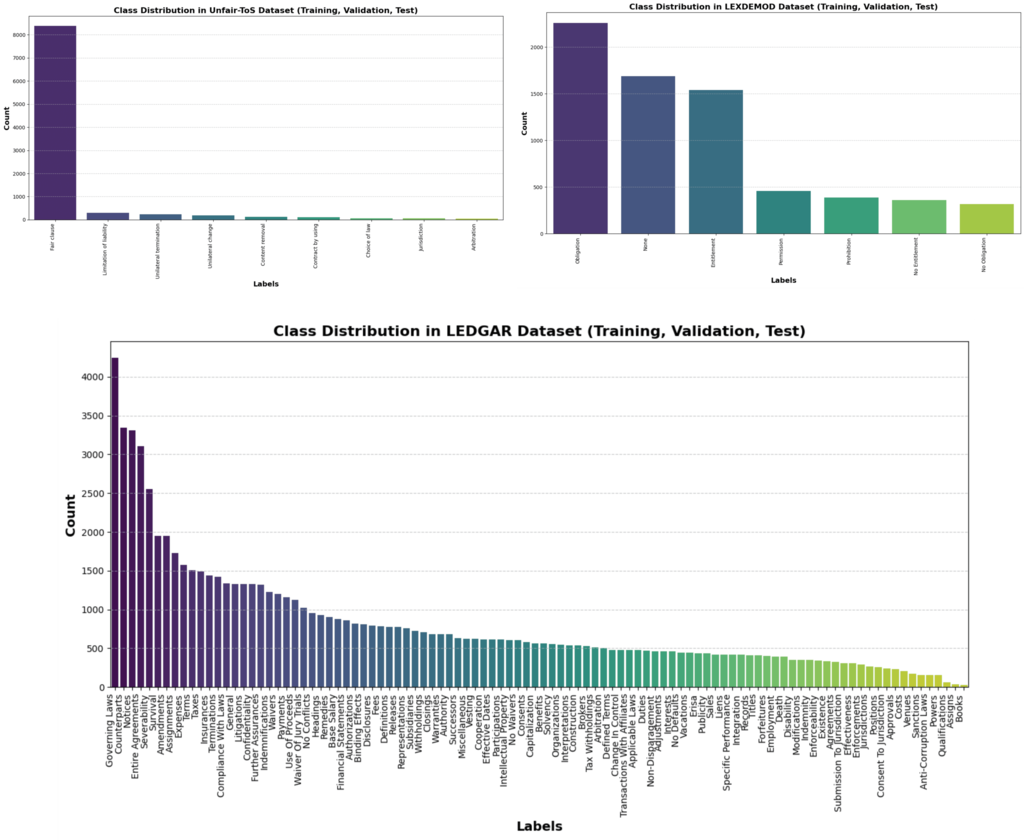}
    \caption{Class Distribution across all three datasets}
    \label{Appendix_Figure0}
\end{figure*}

\begin{table*}[ht!]
\centering
\begin{tabular}{p{2cm} p{2.5cm} p{3.5cm} p{2.2cm} p{2.5cm} p{1cm}} 
  \toprule
  \textbf{Dataset} & \textbf{Contract Type} & \textbf{Task} & \textbf{Task Type} & \textbf{\begin{tabular}[c]{@{}c@{}}Train /Dev/ Test\end{tabular}} & \textbf{Class} \\ \midrule
  UNFAIR-ToS \cite{chalkidis2022lexglue} & Terms of Service (Consumer Contract) & Unfair Contractual Terms Identification & Multi-label Classification & 5,532/2,275/ 1,607 & 9 \\ 
  LEDGAR \cite{chalkidis2022lexglue} & Exhibit-10 Material Contract & Contract Provision Topic Classification & Multi-class Classification & 60,000/10,000/ 10,000 & 100 \\ 
  LEXDEMOD \cite{sancheti2022agent} & Lease Contract & Agent-Specific Deontic Modality Detection & Multi-label Classification & 4,282/330/ 1,777 & 7 \\ \bottomrule
\end{tabular}
\caption{Overview of Datasets used for Benchmarking Legal-specific Models.}
\label{Table1}
\end{table*}
This appendix provides an overview of the datasets used for benchmarking legal-specific models in Table \ref{Table1} and presents labeled examples from all three datasets to aid understanding. The class distribution of all three datasets is given in Figure \ref{Appendix_Figure0}. We provide our own rationales explaining label types, which the datasets do not explicitly include. These explanations clarify why specific labels apply to given clauses (sentences) or provisions (paragraphs), as detailed in Table~\ref{Appendix_Table1}. Legal contract classification involves longer texts than typical NLP tasks like tweets or reviews. Legal-specific Transformer models such as Legal-BERT process up to 512 sub-word tokens, but many LEDGAR paragraphs exceed this limit. Figure~\ref{Appendix_Figure1} shows that numerous LEDGAR paragraphs surpass the standard context window, requiring truncation or other methods to handle long inputs. Additionally, legal texts contain specialized terminology (legalese), increasing classification complexity compared to general text.

\begin{figure*}[ht!]
    \includegraphics[width=1\textwidth]{Appendix_Figure1.pdf}
    \caption{Distribution of text lengths, measured in Legal-BERT subword units, across all three datasets}
    \label{Appendix_Figure1}
\end{figure*}

\begin{table*}[ht!]
\renewcommand{\arraystretch}{1.5}
\Huge
\resizebox{\textwidth}{!}{
\begin{tabular}{clll}
\hline
{\color[HTML]{000000} \textbf{Dataset}} &
  \multicolumn{1}{c}{{\color[HTML]{000000} \textbf{Example}}} &
  \multicolumn{1}{c}{{\color[HTML]{000000} \textbf{Label}}} &
  \multicolumn{1}{c}{{\color[HTML]{000000} \textbf{\begin{tabular}[c]{@{}c@{}}Rationale for Assigned Labels \\ (Provided by us for better understanding)\end{tabular}}}} \\ \hline
{\color[HTML]{CD9934} } &
  {\color[HTML]{000000} \begin{tabular}[c]{@{}l@{}}This Amendment may be executed by one or more of the \\ parties hereto on any number of separate counterparts, \\ and all of said counterparts taken together shall be deemed \\ to constitute one and the same instrument. This \\ Amendment may be delivered by facsimile or other \\ electronic transmission of the relevant signature pages \\ hereof.\end{tabular}} &
  {\color[HTML]{000000} Counterparts} &
  {\color[HTML]{000000} \begin{tabular}[c]{@{}l@{}}This sentence states that the Amendment may be \\ executed in multiple counterparts and that together \\ they form a single agreement, which is a standard \\ counterparts clause used to validate separately \\ signed copies as one binding document.\end{tabular}} \\ \cline{2-4} 
{\color[HTML]{CD9934} } &
  {\color[HTML]{000000} \begin{tabular}[c]{@{}l@{}}THIS AMENDMENT SHALL BE GOVERNED BY, AND \\ INTERPRETED IN ACCORDANCE WITH, THE LAW \\ OF THE STATE OF NEW YORK . The other provisions of \\ Article IX of the Credit Agreement shall apply to this \\ Amendment to the same extent as if fully set forth herein.\end{tabular}} &
  {\color[HTML]{000000} Governing Laws} &
  {\color[HTML]{000000} \begin{tabular}[c]{@{}l@{}}This sentence specifies that New York law will \\ govern and interpret the Amendment, which is a \\ standard governing law clause that establishes \\ the legal jurisdiction and framework for resolving \\ disputes.\end{tabular}} \\ \cline{2-4} 
\multirow{-14}{*}{{\color[HTML]{CD9934} \textbf{LEDGAR}}} &
  {\color[HTML]{000000} \begin{tabular}[c]{@{}l@{}}Sublessee leases the Aircraft in its “as is, where is” condition. \\ The only services, rights, or warranties to which the Sublessee \\ is entitled to under this Sublease are those to which the \\ Sublessor is provided under the Prime Lease.\end{tabular}} &
  {\color[HTML]{000000} Warranties} &
  {\color[HTML]{000000} \begin{tabular}[c]{@{}l@{}}The sentence is labeled as warranties because it \\ defines the rights and guarantees the Sublessee \\ receives and limits those warranties to what the \\ Sublessor has under the Prime Lease.\end{tabular}} \\ \hline
{\color[HTML]{656565} } &
  {\color[HTML]{000000} \begin{tabular}[c]{@{}l@{}}Niantic further reserves the right to remove any User \\ Content from the Service at any time and without notice \\ and for any reason.\end{tabular}} &
  {\color[HTML]{000000} Content removal} &
  {\color[HTML]{000000} \begin{tabular}[c]{@{}l@{}}This sentence is labeled as content removal unfair  \\ contractual term because it gives the provider full \\ control to remove content at any time, for any \\ reason, and without notice.\end{tabular}} \\ \cline{2-4} 
{\color[HTML]{656565} } &
  {\color[HTML]{000000} \begin{tabular}[c]{@{}l@{}}amazon reserves the right to refuse service, terminate \\ accounts , terminate your rights to use amazon services, \\ remove or edit content , or cancel orders in its sole \\ discretion.\end{tabular}} &
  {\color[HTML]{000000} \begin{tabular}[c]{@{}l@{}}Unilateral \\ termination,\\ Content removal\end{tabular}} &
  {\color[HTML]{000000} \begin{tabular}[c]{@{}l@{}}This sentence is labeled as unilateral termination and \\ content removal because it allows Amazon to end \\ services, remove content, or cancel orders at its sole \\ discretion, without notice, creating an imbalance of \\ power.\end{tabular}} \\ \cline{2-4} 
\multirow{-10}{*}{{\color[HTML]{656565} \textbf{UNFAIR-ToS}}} &
  {\color[HTML]{000000} \begin{tabular}[c]{@{}l@{}}Outside the United States and Canada. If you acquired \\ the if you acquired the application in any other country, \\ the laws of that country apply.\end{tabular}} &
  {\color[HTML]{000000} None} &
  {\color[HTML]{000000} \begin{tabular}[c]{@{}l@{}}The sentence is labeled as none because it does not \\ belong to any of the unfair contractual term types \\ and is actually a fair clause.\end{tabular}} \\ \hline
{\color[HTML]{6665CD} } &
  {\color[HTML]{000000} \begin{tabular}[c]{@{}l@{}}{[}lessee{]} Lessee will not create or permit to be \\ created or to remain , and will promptly \\ discharge , any lien , encumbrance or charge (including \\ without limitation any mechanic 's , laborer 's or \\ materialman 's lien ) against the Premises or any part \\ thereof arising from Lessee 's actions.\end{tabular}} &
  {\color[HTML]{000000} \begin{tabular}[c]{@{}l@{}}Prohibition, \\ Obligation\end{tabular}} &
  {\color[HTML]{000000} \begin{tabular}[c]{@{}l@{}}This sentence imposes a prohibition by forbidding \\ the Lessee (the agent) from creating or allowing liens. \\ It also imposes an obligation by requiring the Lessee \\ to promptly remove any such liens.\end{tabular}} \\ \cline{2-4} 
{\color[HTML]{6665CD} } &
  {\color[HTML]{000000} \begin{tabular}[c]{@{}l@{}}{[}tenant{]} Tenant may, without Landlord's consent, \\ before delinquency occurs, contest any such taxes related \\ to the Personal Property.\end{tabular}} &
  {\color[HTML]{000000} Permission} &
  {\color[HTML]{000000} \begin{tabular}[c]{@{}l@{}}This sentence grants permission to the Tenant (the \\ agent) to contest taxes without needing the \\ Landlord’s consent, as long as it's done before \\ delinquency.\end{tabular}} \\ \cline{2-4} 
\multirow{-12}{*}{{\color[HTML]{6665CD} \textbf{LEXDEMOD}}} &
  {\color[HTML]{000000} \begin{tabular}[c]{@{}l@{}}{[}landlord{]} Tenant shall promptly notify Landlord of any \\ alleged defaults under the CC\&Rs and/or the Oil and Gas \\ Lease .\end{tabular}} &
  {\color[HTML]{000000} Entitlement} &
  {\color[HTML]{000000} \begin{tabular}[c]{@{}l@{}}The Landlord, as the agent, holds an entitlement to\\ receive notice from the Tenant about alleged defaults.\end{tabular}} \\ \hline
\end{tabular}
}
\caption{Overview of all three Datasets with Examples, Labels, and Author-Provided Rationales}
\label{Appendix_Table1}
\end{table*}

\section{Description of Legal-specific Models}
\label{AppendixB}
\begin{table*}[ht!]
\centering
\resizebox{0.9\textwidth}{!}{
\begin{tabular}{llll}
\toprule
\textbf{Legal-Specific Model} &
  \textbf{Pre-training Corpora} &
  \textbf{\# Doc} &
  \textbf{Base Model} \\
\midrule
\begin{tabular}[c]{@{}l@{}}Legal-BERT \\ \cite{chalkidis2020legal}\end{tabular} &
  \begin{tabular}[c]{@{}l@{}}\href{https://eur-lex.europa.eu/}{EU Legislation}, \href{https://www.legislation.gov.uk/}{UK Legislation}, \href{https://eur-lex.europa.eu/}{European Court of Justice (ECJ)}\\ \href{https://eur-lex.europa.eu/}{Cases}, \href{https://hudoc.echr.coe.int/}{European Court of Human Right (ECHR) Cases}, \href{https://case.law/}{US Court}\\ \href{https://case.law/}{Cases}, \href{https://www.sec.gov/edgar/search/}{US Contracts}\end{tabular} &
  354K &
  \texttt{BERT-base-uncased} \\
\begin{tabular}[c]{@{}l@{}}Contracts-BERT \\ \cite{chalkidis2020legal}\end{tabular} &
    \href{https://www.sec.gov/edgar/search/}{US Contracts} &
  76K &
  \texttt{BERT-base-uncased} \\
\begin{tabular}[c]{@{}l@{}}Legal-RoBERTa \\ \cite{geng2021legal}\end{tabular} &
  \begin{tabular}[c]{@{}l@{}}\href{https://www.kaggle.com/datasets/uspto/patent-litigations}{Patent Litigations}, \href{https://case.law/}{US Court Cases}, \href{https://www.kaggle.com/datasets/bigquery/patents}{Google Patents Public Data}\end{tabular} &
  - &
  \texttt{RoBERTa-base} \\
\begin{tabular}[c]{@{}l@{}}CaseLaw-BERT \\ \cite{zheng2021does}\end{tabular} &
  \begin{tabular}[c]{@{}l@{}}\href{https://case.law/}{Harvard Case Law (US federal and State courts)}\end{tabular} &
  3.4M &
  \texttt{BERT-base-uncased} \\
\begin{tabular}[c]{@{}l@{}}PoL-BERT \\ \cite{henderson2022pile}\end{tabular} &
  \begin{tabular}[c]{@{}l@{}}\href{https://huggingface.co/datasets/pile-of-law/pile-of-law}{Court Opinions, Government, Publications, Contracts, Statutes,}\\ \href{https://huggingface.co/datasets/pile-of-law/pile-of-law}{Legal Analyses, Regulations, and, more from US and EU}\end{tabular} &
  10M &
  \texttt{RoBERTa-large} \\
\begin{tabular}[c]{@{}l@{}}InLegalBERT \\ \cite{paul-2022-pretraining}\end{tabular} &
  \begin{tabular}[c]{@{}l@{}}\href{https://www.sci.gov.in/}{Indian Supreme Court}, \href{https://indiankanoon.org/}{High Court, and District Court Cases,} \\ \href{https://indiankanoon.org/}{Central Government Acts of India}\end{tabular} &
  5.4M &
  \texttt{Legal-BERT-base-uncased} \\
\begin{tabular}[c]{@{}l@{}}InCaseLawBERT \\ \cite{paul-2022-pretraining}\end{tabular} &
  \begin{tabular}[c]{@{}l@{}}\href{https://www.sci.gov.in/}{Indian Supreme Court}, \href{https://indiankanoon.org/}{High Court, and District Court Cases,} \\ \href{https://indiankanoon.org/}{Central Government Acts of India}\end{tabular} &
  5.4M &
  \texttt{CaseLaw-BERT-base-uncased} \\
\begin{tabular}[c]{@{}l@{}}CustomInLawBERT \\ \cite{paul-2022-pretraining}\end{tabular} &
  \begin{tabular}[c]{@{}l@{}}\href{https://www.sci.gov.in/}{Indian Supreme Court}, \href{https://indiankanoon.org/}{High Court, and District Court Cases,} \\ \href{https://indiankanoon.org/}{Central Government Acts of India}\end{tabular} &
  5.4M &
  \texttt{BERT-base-uncased} \\
\begin{tabular}[c]{@{}l@{}}LexLM \\ \cite{chalkidis-garneau-etal-2023-lexlms}\end{tabular} &
  \begin{tabular}[c]{@{}l@{}}\href{https://huggingface.co/datasets/lexlms/lex\_files}{EU Legislation and Case Law, UK Legislation and Case Law,} \\ \href{https://huggingface.co/datasets/lexlms/lex\_files}{Canadian Legislation and Case Law, U.S. Case Law and} \\ \href{https://huggingface.co/datasets/lexlms/lex\_files}{Contracts, ECHR Case Law, and Indian Case Law}\end{tabular} &
  5.8M &
  \texttt{RoBERTa-base} \\
\begin{tabular}[c]{@{}l@{}}Legal-XLM-R \\ \cite{niklaus2024multilegalpile}\end{tabular} &
  \begin{tabular}[c]{@{}l@{}}\href{https://huggingface.co/datasets/joelniklaus/Multi\_Legal\_Pile}{Different Countries Case laws and legislation, US/EU contracts, }\\ \href{https://huggingface.co/datasets/joelniklaus/Multi\_Legal\_Pile}{and other legal-specific documents}\end{tabular} &
  59M &
  \texttt{XLM-RoBERTa-base} \\
\begin{tabular}[c]{@{}l@{}}LexT5 \\ \cite{t-y-s-s-etal-2024-lexsumm}\end{tabular} &
  \begin{tabular}[c]{@{}l@{}}\href{https://huggingface.co/datasets/lexlms/lex\_files}{EU Legislation and Case Law, UK Legislation and Case Law,} \\ \href{https://huggingface.co/datasets/lexlms/lex\_files}{Canadian Legislation and Case Law, U.S. Case Law and} \\ \href{https://huggingface.co/datasets/lexlms/lex\_files}{Contracts, ECHR Case Law, and Indian Case Law}\end{tabular} &
  5.8M &
  \texttt{T5-base} \\
\begin{tabular}[c]{@{}l@{}}AdaptLLM \\ \cite{cheng2024adapting}\end{tabular} &
  \begin{tabular}[c]{@{}l@{}}\href{https://huggingface.co/datasets/AdaptLLM/law-tasks}{US court opinions from FreeLaw project}\end{tabular} &
  3.6M &
  \texttt{LLama-1} \\
\begin{tabular}[c]{@{}l@{}}SaulLM-7B \\ \cite{cheng2024adapting}\end{tabular} &
  \begin{tabular}[c]{@{}l@{}}\href{https://huggingface.co/Equall}{FreeLaw, English MultiLegal Pile, EDGAR, EuroParl, GovInfo}\\ \href{https://huggingface.co/Equall}{Law Stack Exchange, Australian legal corpora, EU and UK}\\\href{https://huggingface.co/Equall}{legislation, court transcripts, and USPTO filings}  \end{tabular} & - &
  \texttt{Mistral} \\
\bottomrule
\end{tabular}
}
\caption{Key specifications of the evaluated models, including pre-training corpora (with links), document counts, and base models used.}
\label{Table2}
\end{table*}
Table \ref{Table2} summarizes their key characteristic. The detailed description of the legal-specific models used in the experiments is given below:

\textbf{Legal-BERT} Legal-BERT \cite{chalkidis2020legal} is a BERT-base-uncased model (110M parameters) pre-trained on 354K English legal documents, including EU and UK legislation, US contracts, and US and EU court cases. It follows the original BERT pre-training configuration and constructs its sub-word vocabulary from scratch to better capture legal terminology.

\textbf{Contracts-BERT} Contracts-BERT \cite{chalkidis2020legal} is a BERT-base-uncased model (110M parameters) pre-trained on 76K US contracts. It follows the original BERT configuration and retains a custom vocabulary tailored to contract language. 

\textbf{Legal-RoBERTa} Legal-RoBERTa \cite{geng2021legal} builds on the RoBERTa-base model (125M parameters) and continues pre-training on 4.9 GB of legal text, including patent litigation documents, US court cases, and publicly available Google Patents data. 

\textbf{CaseLaw-BERT} CaseLaw-BERT \cite{zheng2021does} is a BERT-base-uncased model (110M parameters) pre-trained on 3.4M US federal and state court decisions from the Harvard Case Law corpus. Although originally referred to as \emph{Custom Legal-BERT} by \cite{zheng2021does}, it is later termed \emph{CaseLaw-BERT} by \cite{chalkidis2022lexglue} to distinguish it from the earlier Legal-BERT of \cite{chalkidis2020legal}, highlighting its exclusive training on harvard case law. This naming convention is now widely adopted, and we follow the same in this work. 

\textbf{PoL-BERT} PoL-BERT \cite{henderson2022pile} is a RoBERTa-large model (340M parameters) pre-trained on the \emph{Pile-of-Law}, a 256GB corpus comprising 10M legal and administrative documents. The dataset spans a wide range of legal domains, including US federal and state court opinions (e.g., CourtListener, SCOTUS filings), regulatory documents (e.g., Federal Register, Code of Federal Regulations, SEC and IRS guidance), legislative texts (e.g., US Bills, US Code, State Codes), and other legal document sources (e.g., ECHR, Eur-Lex, ICJ/PCIJ rulings). It also includes administrative decisions from US agencies (e.g., DOJ, OLC, BVA, NLRB, EOIR, DOL), legal contracts (e.g., EDGAR filings, Atticus contracts, CFPB agreements), educational materials (e.g., open-access casebooks, exam outlines), and publicly available community-driven legal discussions.

\textbf{InLegalBERT}
InLegalBERT \cite{paul-2022-pretraining} builds on Legal-BERT-base-uncased \cite{chalkidis2020legal}, a legal-specific BERT model (110M parameters) initially pre-trained on 354K English legal documents, including EU and UK legislation, US contracts, and US and EU court cases. It is further pre-trained on 5.4M Indian legal documents, including judgments from the Supreme Court, High Courts, and District Courts, as well as Central Government Acts of India. This extended pre-training enables the model to better capture the linguistic and legal nuances of other jurisdictions, such as Indian jurisprudence. 

\textbf{InCaseLawBERT}
InCaseLawBERT \cite{paul-2022-pretraining} builds on CaseLaw-BERT-base-uncased (110M parameters), which is initially pre-trained on 3.4M US federal and state court decisions from the Harvard Case Law corpus. It undergoes further pre-training on 5.4M Indian legal documents, including judgments from the Supreme Court, High Courts, and District Courts, as well as Central Government Acts of India. This additional training enables the model to better capture the linguistic and legal nuances of other jurisdictions, particularly Indian jurisprudence. 

\textbf{CustomInLawBERT}
CustomInLawBERT \cite{paul-2022-pretraining} is a BERT-base-uncased model (110M parameters) pre-trained from scratch on 5.4M Indian legal documents, including judgments from the Supreme Court, High Courts, and District Courts, as well as Central Government Acts of India. 

\textbf{LexLMs}
LexLMs \cite{chalkidis-garneau-etal-2023-lexlms} include two variants: RoBERTa-base (124M parameters) and RoBERTa-large (340M parameters), both pre-trained from scratch on 5.8M legal documents from multiple English-speaking jurisdictions. The corpus covers a wide range of sources, including EU legislation, EU and ECtHR court decisions, UK legislation and court cases, Indian court decisions, Canadian legislation and court decisions, US court decisions, US legislation, and US contracts. The dataset is designed to ensure broad jurisdictional and document-type coverage, with US legal texts comprising the largest portion. This large-scale, English legal-domain pre-training enables LexLMs to support robust legal language understanding across common law and mixed legal systems.

\textbf{Legal-XLM-R}
Legal-XLM-R \cite{niklaus2024multilegalpile} includes two variants: RoBERTa-base (124M parameters) and RoBERTa-large (340M parameters), both pre-trained from scratch on a multilingual legal corpus comprising 59M documents. The dataset spans 24 languages and five legal text types, including legislation and case law, collected from various jurisdictions such as Germany, Switzerland, the UK, and several other countries. This large-scale, cross-lingual pre-training enables Legal-XLM-R to support legal language understanding across multilingual and multi-jurisdictional contexts.

\textbf{LexT5} LexT5 \cite{t-y-s-s-etal-2024-lexsumm} is a legal-oriented sequence-to-sequence model designed to address the limitations of encoder-only architectures in legal NLP. It is pre-trained on three T5 variants, T5 Small (60M parameters), T5 Base (220M), and T5 Large (770M), using the same 5.8 million legal documents employed for LexLMs \cite{chalkidis-garneau-etal-2023-lexlms}.

\textbf{AdaptLLM}
AdaptLLM \cite{cheng2024adapting} is a LLaMA-1 model (7B parameters) pre-trained on 3.6M US court opinions from the FreeLaw project \cite{gao2020pile}. It converts raw legal texts into reading comprehension tasks, such as summarization, NLI, commonsense reasoning, and text completion, using regex-based patterns, and learns in a self-supervised manner. 
\begin{figure*}[!t]
    \centering
    \includegraphics[width=0.8\textwidth]{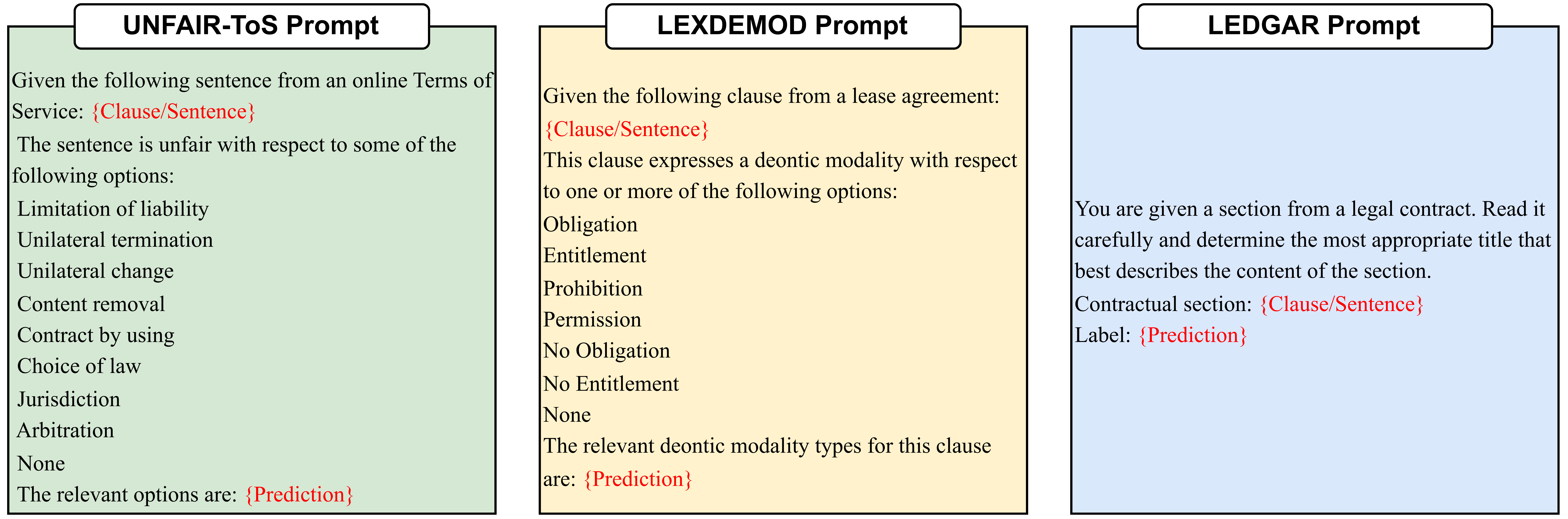}
    \caption{Prompt Templates for Instruction-Based Fine-Tuning of LexT5}
    \label{Appendix_Figure2}
\end{figure*}
\textbf{SaulLM-7B}
SaulLM-7B \cite{colombo2024saullm} is a Mistral-based model (7B parameters) pre-trained on 94B tokens of legal text. It combines data from existing sources such as FreeLaw, the English MultiLegal Pile, EDGAR, and EuroParl, along with additional datasets including GovInfo, Law Stack Exchange, Australian legal corpora, EU and UK legislation, court transcripts, and USPTO filings. The corpus is curated from both public datasets and web-scraped legal resources to ensure broad legal coverage across jurisdictions.

\section{Experimental Setup}
\label{AppendixC}
We use all publicly available legal-specific pre-trained models from Hugging Face. To ensure fair comparison, we adopt the training configuration introduced by \citet{chalkidis2022lexglue} for the LEDGAR and UNFAIR-ToS datasets: a learning rate of $3\text{e-}5$ for all nine encoder-base models and $1\text{e-}5$ for the encoder-large model PoL-BERT \cite{henderson2022pile}, consistent with the setting used for \texttt{RoBERTa-large}. All models are trained for up to 20 epochs with a batch size of 8, using early stopping with a patience of 3 based on development set performance. For UNFAIR-ToS, we use a maximum sequence length of 128, as in \citet{chalkidis2022lexglue}. However, we disable mixed-precision training (i.e., set fp16=False) to ensure stable training, which results in longer training times compared to \citet{chalkidis2022lexglue}. For LEDGAR, we reduce the maximum sequence length to 128 (from 512 used in \citet{chalkidis2022lexglue}) to save computational resources and training time. This adjustment is necessary because, as discussed above, fp16 is disabled to ensure stable training, which leads to longer training times, and the LEDGAR dataset contains over 80k sentences, which is large. We observe only a marginal performance drop (0.1-0.4\%), which we consider acceptable for efficiency. Each model is trained five times with different random seeds (1-5), and we report the test results of the best seed, following \citet{chalkidis2022lexglue} for a fair baseline comparison. For the LEXDEMOD dataset, we follow the setup proposed by \citet{sancheti2022agent}, using a learning rate of $2\text{e-}5$ for all encoder-based legal-specific models, including PoL-BERT, consistent with their configuration for \texttt{RoBERTa-large}. We use a batch size of 8 and apply early stopping with a patience of 3. The maximum sequence length is set to 256, as in \citet{sancheti2022agent}. Each model is trained five times with different random seeds, and we report the average test performance across the three best seeds, following \citet{sancheti2022agent} for a fair baseline comparison. For decoder-based generalist and legal-specific models, the parameters remain the same. However, for the LEDGAR dataset, the number of epochs for models such as Mistral, AdaptLLM, and SaulLM is reduced to 3 instead of 20 due to the long training time (22-24 hours per model). This adjustment is made because fp16 is disabled for stable training, and it is necessary for LEDGAR due to its large model size and the resource constraints. In addition to encoder-decoder based legal-specific model (LexT5), we adopt instruction-based fine-tuning, which aligns better with encoder-decoder models \cite{wang2023aligning}. This approach pairs natural language prompts with clause inputs, enabling the model to generate the appropriate label(s) as output. The instruction templates used are listed in Figure \ref{Appendix_Figure2}. For UNFAIR-ToS, we directly use the prompt from \citet{chalkidis2023chatgpt}, originally designed for zero-shot prompting. For LEXDEMOD and LEDGAR, we design our own prompts inspired by that style. We evaluate model performance using micro-F1 ($\mu$-F1) and macro-F1 (m-F1) to account for class imbalance. Additionally, we report the arithmetic mean with standard deviation for micro-F1 ($\mu$-F1) and macro-F1 (m-F1) across tasks. All experiments are conducted on a single NVIDIA V100 GPU. 
\section{Other generalist Models Considered for Exploratory Purposes}
\label{AppendixE}
 Here, we discuss the closed-source model (GPT-3.5-Turbo) results using prompting techniques, as reported by \cite{chalkidis2023chatgpt}. For UNFAIR-ToS, zero-shot prompting yields micro and macro F1-scores of 41.4 and 22.2, respectively. In few-shot prompting, performance improves, achieving micro and macro F1-scores of 64.7 and 32.5, though these scores remain lower than those achieved through legal-specific and generalist model fine-tuning techniques. A similar pattern occurs with LEDGAR, where zero-shot prompting results in micro and macro F1-scores of 70.1 and 56.7, respectively, while few-shot prompting produces scores of 62.1 and 51.1. Here, performance decreases with few-shot prompting, but task-specific fine-tuning using legal and generalist models remains the preferred approach. Thus, task-specific fine-tuning in the legal domain is essential for accurate legal contract classification.
\end{document}